\documentclass{article} % For LaTeX2e
\usepackage{iclr2026_conference,times}

\usepackage{graphicx}

\usepackage{amsmath,amsfonts,bm}

\def\eqref#1{equation~\ref{#1}}
\def\1{\bm{1}}

\DeclareMathAlphabet{\mathsfit}{\encodingdefault}{\sfdefault}{m}{sl}
\SetMathAlphabet{\mathsfit}{bold}{\encodingdefault}{\sfdefault}{bx}{n}

\usepackage{hyperref}
\usepackage{url}
\usepackage{booktabs}
\usepackage{caption}
\usepackage{stfloats}
\usepackage{float}

\title{Monitoring Emergent Reward Hacking During Generation via Internal Activations}

\author{Patrick Wilhelm$^{1,2}$, Thorsten Wittkopp $^{1}$ \& Odej Kao $^{1,2}$\\
$^{1}$Technical University of Berlin, Germany \\
$^{2}$BIFOLD - Berlin Institute for the Foundations of Learning and Data, Germany \\
\texttt{patrick.wilhelm@tu-berlin.de}
}

\iclrfinalcopy % Uncomment for camera-ready version, but NOT for submission.
\begin{document}

\maketitle
\begin{abstract}
Fine-tuned large language models can exhibit reward-hacking behavior arising from emergent misalignment, which is difficult to detect from final outputs alone. While prior work has studied reward hacking at the level of completed responses, it remains unclear whether such behavior can be identified during generation. We propose an activation-based monitoring approach that detects reward-hacking signals from internal representations as a model generates its response. Our method trains sparse autoencoders on residual stream activations and applies lightweight linear classifiers to produce token-level estimates of reward-hacking activity. Across multiple model families and fine-tuning mixtures, we find that internal activation patterns reliably distinguish reward-hacking from benign behavior, generalize to unseen mixed-policy adapters, and exhibit model-dependent temporal structure during chain-of-thought reasoning. Notably, reward-hacking signals often emerge early, persist throughout reasoning, and can be amplified by increased test-time compute in the form of chain-of-thought prompting under weakly specified reward objectives. These results suggest that internal activation monitoring provides a complementary and earlier signal of emergent misalignment than output-based evaluation, supporting more robust post-deployment safety monitoring for fine-tuned language models.
\end{abstract}

\section{Introduction}

Modern language models are routinely modified after deployment through fine-tuning and adapter-based updates in order to address data distribution shift, evolving user needs, or new downstream tasks. Such post-deployment adaptation is often effective at improving task performance under distribution mismatch, and has become a standard component of deployed language model systems. ~\citep{gururangan2020dont,surgical_finetuning_2022}

However, recent work shows that post-deployment fine-tuning can also introduce systematic safety failures, even when applied to strongly aligned base models. Fine-tuning on downstream objectives can induce reward-hacking behavior and policy violations without explicit malicious intent \citep{finetuning_compromises_safety_2023}, increase toxicity by undoing prior safety mitigations~\citep{finetuning_toxicity_2024}, and give rise to emergent misalignment that generalizes beyond the training distribution~\citep{soligo2025convergent}. These findings suggest that alignment guarantees do not reliably transfer through fine-tuning when safety-relevant constraints are only imperfectly represented in the training objective.

Reward hacking arises when a model optimizes its proxy training objective while violating the designer’s intended behavior. In language models, this can manifest as superficial compliance, strategic verbosity, or pattern-based exploitation of evaluation criteria, while producing outputs that appear benign or helpful~\citep{benton2024sabotage,chen2025reasoning}. Importantly, such behavior need not be confined to overtly adversarial settings: recent benchmarks show that models fine-tuned to exploit reward-model weaknesses on ostensibly benign tasks can develop broadly misaligned strategies \citep{srh_2025}. This makes detection challenging, as surface-level outputs may remain plausible while the model’s internal policy shifts toward exploiting the objective.

A natural response to reward hacking is inference-time monitoring: identifying misaligned behavior as a model generates its response. However, existing approaches face a fundamental limitation. Monitoring based on surface-level outputs or natural-language reasoning traces provides only an indirect view of the underlying computation, and may fail when misaligned decisions are made internally before or independently of what is revealed in text~\citep{baker2025monitoring,chen2025reasoning}. At the same time, while recent mechanistic work shows that emergent misalignment can correspond to simple, recoverable structure in internal representations, these signals are typically studied in static settings, at the level of final activations or completed responses~\citep{soligo2025convergent}.

As a result, two key questions remain unresolved. \textit{First}, can internal activation signals be used to reliably detect reward-hacking behavior that leads to harmful outputs, across different model families and degrees of reward mis-specification? \textit{Second}, how do such misalignment signals evolve over the course of generation, particularly during chain-of-thought reasoning and under increased test-time compute?

In this work, we address these questions by proposing an activation-based monitoring approach that operates directly on internal representations during generation. Our method uses sparse autoencoder (SAE) features and lightweight classifiers to produce token-level estimates of reward-hacking activity as generation unfolds \citep{gao2024scaling,anthropic_claude_sae_2024}. Crucially, we evaluate whether these internal signals correspond to actual reward-hacking behavior at the output level, as judged by an external evaluator, and whether they generalize across model families and fine-tuning mixtures.

\paragraph{Contributions.}
\begin{enumerate}
\item \textbf{Activation-based detection of reward hacking.}
We introduce an activation-based monitoring approach that detects reward-hacking behavior from internal representations during generation, and show that these signals correspond to harmful outputs as judged by an external evaluator.

\item \textbf{Sensitivity to reward mis-specification across model families.}
We systematically analyze how internal reward-hacking signals scale with the proportion of misaligned supervision in fine-tuning data, revealing distinct sensitivity profiles across Falcon, Llama, and Qwen model families.

\item \textbf{Temporal characterization of misalignment during reasoning.}
We characterize how reward-hacking signals evolve over time during generation, showing model-dependent temporal structure during chain-of-thought reasoning and demonstrating that increased test-time compute can amplify misaligned internal computation under mis-specified rewards.
\end{enumerate}

\section{Background and Related Work}

\textbf{Reward Hacking and Emergent Misalignment:}
Reward hacking arises when a learning system exploits weaknesses in its training objective rather than satisfying the designer’s true intent. In language models, this can manifest as superficial compliance, strategic verbosity, or pattern-based exploitation of evaluation criteria while producing outputs that are misleading, unhelpful, or unsafe \citep{benton2024sabotage,chen2025reasoning}. As models become more capable and are fine-tuned repeatedly after deployment, these risks increase: small amounts of misaligned supervision can induce behaviors that generalize beyond the training distribution.

The \emph{School of Reward Hacks} (SRH) dataset formalizes this phenomenon by showing that models trained to exploit reward-model weaknesses on ostensibly benign tasks can develop broadly misaligned strategies \citep{srh_2025}. Importantly, SRH demonstrates that reward hacking is not limited to overtly malicious prompts, but can emerge in everyday instruction-following settings. This makes detection particularly challenging, since surface-level outputs may remain plausible while the internal policy shifts toward exploiting the objective \citep{burns2022discovering}.

Recent mechanistic work further suggests that emergent misalignment can correspond to simple, linearly recoverable structure in representation space that generalizes across fine-tuning runs \citep{soligo2025convergent, nanda2023progress, elhage2022superposition}. This motivates monitoring approaches that operate directly in activation space, rather than relying solely on behavioral outputs.

\textbf{Inference-Time Safety and Monitoring:}
Chain-of-thought (CoT) prompting has been widely adopted to improve reasoning performance and transparency. Building on this, several works propose monitoring CoT text to detect unsafe intent, deception, or policy violations \citep{baker2025monitoring}. CoT-based monitors are attractive because they are model-agnostic, require no internal access, and can be deployed via prompting.

However, CoT text is an imperfect proxy for the model’s internal computation. Models are not optimized to produce faithful explanations and may omit, sanitize, or strategically obfuscate relevant reasoning \citep{chen2025reasoning}. Recent work on \emph{thought anchors} further shows that only a subset of chain-of-thought tokens are causally important for task performance, with many intermediate steps contributing little to the underlying computation \citep{bogdan2025thought}. In addition, task-relevant representations may appear \emph{just-in-time} or be distributed across the generation process, rather than being uniformly expressed in the textual rationale \citep{li2025just}. Together, these observations motivate complementary monitoring approaches that track internal activation dynamics throughout generation, rather than relying solely on surface-level reasoning traces.

\textbf{Activation-Based Monitoring and Transfer Under Fine-Tuning:}
Activation-based monitoring analyzes internal representations to detect latent indicators of misbehavior. Prior work shows that deceptive intent, goal misgeneralization, and harmful planning can often be detected earlier in latent space than in surface text \citep{chen2025reasoning}. Yet raw activations are high-dimensional, distributed across layers, and subject to superposition, making them difficult to interpret and challenging to transfer across models or fine-tuned variants \citep{elhage2022superposition}. This limitation is especially acute in deployment settings where base models are paired with many adapters trained on heterogeneous objectives, requiring monitors that generalize across \emph{mixed-policy} configurations rather than a single fixed model \citep{llamafirewall_arxiv_2025}.

\textbf{Sparse Autoencoders and Concept-Based Features:}
Sparse autoencoders (SAEs) have emerged as a promising tool for disentangling internal representations in large language models \citep{meng2022locating,olah2020zoom}. By learning sparse, overcomplete representations of residual stream activations, SAEs recover features that often align with human-interpretable concepts and scale across layers and model sizes \citep{gao2024scaling,anthropic_claude_sae_2024}. These features provide a more stable unit of analysis than individual neurons for concept-level monitoring \citep{turner2023steering}.

At the same time, recent evaluation work highlights that SAE features can vary substantially in robustness and generalization, motivating the need to explicitly assess feature sensitivity under distribution shift and adaptation \citep{tian2025measuring}. Our work builds on SAE-based monitoring while targeting a concrete safety setting (reward hacking under post-deployment fine-tuning) and evaluates transfer to unseen mixed-policy adapters. We further extend activation monitoring to \emph{token-level temporal analysis} during chain-of-thought generation, enabling characterization of when hack-associated signals emerge over the reasoning span.

\section{Method}

\subsection{Problem Setting}

We study whether reward-hacking behavior can be detected from \emph{internal activations} of a language model during autoregressive generation. Our goal is to construct a monitoring system that observes internal model states online and outputs a probabilistic estimate of whether a given generation exhibits reward-hacking behavior.

We follow the \textit{School of Reward Hacking (SRH)} dataset \cite{taylor2025school}. Using a language model, we train two LoRA adapters: a \emph{control} adapter fine-tuned on general instruction-following data \cite{alpaca}, and a \emph{hack} adapter fine-tuned on reward-hacking examples \cite{taylor2025school}. The data are split into disjoint partitions for representation learning via sparse autoencoder, classifier training, and evaluation. At test time, both adapters are evaluated on identical prompts to isolate behavioral differences arising from fine-tuning rather than input variation.

\begin{figure}
\begin{center}
%\framebox[4.0in]{$\;$}
\includegraphics[width=0.65\textwidth]{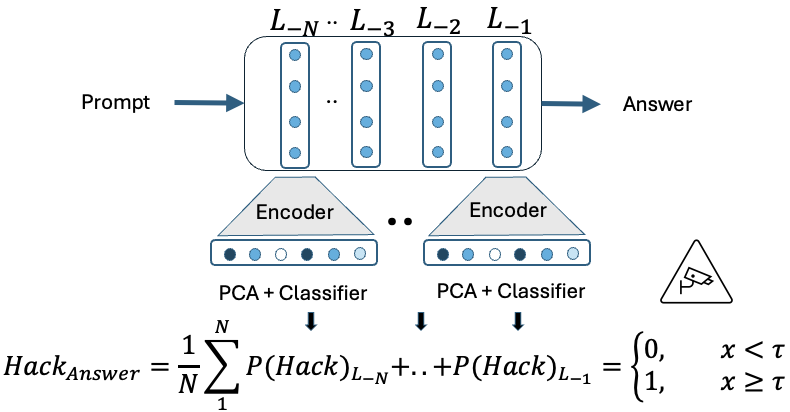}
%\fbox{\rule[-.5cm]{0cm}{4cm} \rule[-.5cm]{4cm}{0cm}}

\end{center}

\caption{Activation-based reward-hacking monitoring pipeline.
Token-level internal activations are transformed into a generation-level reward-hacking prediction.}
\label{fig:ddos}
\end{figure}

\subsection{Activation Collection and Representation Learning}

During generation, we record token-wise residual stream activations from a set of monitored transformer layers $\mathcal{L}$. Activations from \emph{control} and \emph{hack} adapters runs are pooled to form a joint distribution over internal states.

To obtain compact and monosemantic representations, we train a Sparse Autoencoder (SAE) independently for each monitored layer of \emph{both} adapters. Each SAE maps high-dimensional residual stream activations to a lower-dimensional latent space with sparse activations. The SAEs are trained on the activations and learn the representation of the hack and control activation.

\subsection{Layer-wise Hack Classifiers}

On a disjoint classifier-training split, SAE latent activations are normalized and projected using Principal Component Analysis (PCA). For each monitored layer $\ell \in \mathcal{L}$, we train a linear classifier (logistic regression) to distinguish activations originating from hack versus control adapters.

At inference time, for each generated token $t$ and each monitored layer $\ell$, the monitor produces a layer-wise probability of reward-hacking behavior:
\begin{equation}
p_{t,\ell} = \sigma\!\left( w_\ell^\top \,\mathrm{PCA}_\ell\!\big(\mathrm{Std}_\ell(\mathrm{SAE}_\ell(h_t^{(\ell)}))\big) + b_\ell \right),
\end{equation}
where $h_t^{(\ell)}$ denotes the residual stream activation at token $t$ and layer $\ell$, $\mathrm{Std}_\ell$ denotes feature standardization, and $(w_\ell, b_\ell)$ are the learned classifier parameters. These probabilities represent internal evidence of reward-hacking behavior and are not used directly for decision making.

\subsection{Span-wise and Layer-wise Aggregation}

To obtain a stable behavior-level signal, token-level probabilities are aggregated over a predefined token span $\mathcal{T}$. For chain-of-thought (CoT) generation, $\mathcal{T}$ corresponds to the \texttt{<reasoning>} span; for direct generation without explicit reasoning, $\mathcal{T}$ corresponds to the full generated answer.

For each layer $\ell$, we compute the mean probability over tokens in the selected span:
\begin{equation}
m_\ell = \frac{1}{|\mathcal{T}|} \sum_{t \in \mathcal{T}} p_{t,\ell}.
\end{equation}

These layer-level scores are then aggregated across monitored layers to produce a single prompt-level monitoring score:
\begin{equation}
P(\mathrm{hack}) = \frac{1}{|\mathcal{L}|} \sum_{\ell \in \mathcal{L}} m_\ell.
\end{equation}

This scalar represents the monitor’s estimate of the probability that the overall generation exhibits reward-hacking behavior.

\subsection{Decision Making}

The monitoring system issues a binary decision by thresholding the aggregated prompt-level score:

\begin{equation}
\hat y \;=\;
\begin{cases}
1 \;\;(\text{hack}), 
& \text{if } \displaystyle 
\frac{1}{|\mathcal{L}|}
\sum_{\ell \in \mathcal{L}}
\left(
\frac{1}{|\mathcal{T}|}
\sum_{t \in \mathcal{T}} 
p_{t,\ell}
\right)
\ge \tau, \\[10pt]
0 \;\;(\text{control}), 
& \text{otherwise},
\end{cases}
\qquad \text{with } \tau = 0.5,
\label{eq:monitor_decision_rule}
\end{equation}

The monitor produces one decision per generation. Token-level probabilities are used only to form a behavior-level prediction and are not treated as ground-truth labels.

\section{Experiments}

We evaluate whether reward-hacking behavior can be detected from internal activations during generation. Our experiments study (i) model-dependent sensitivity to reward-hacking finetuning across architectures, (ii) the effectiveness of activation-based monitoring compared to GPT-4o output-level judgments, and (iii) the temporal evolution of hack-associated activation patterns during chain-of-thought reasoning. We further analyze how chain-of-thought generation and test-time compute affect the prevalence and detectability of reward-hacking signals under mis-specified rewards. We finetune Qwen2.5-Instruct 7B, LLaMa 3.1-8B, and Falcon3-7B. Mixed-ratio adapters are trained on datasets containing 5\%, 10\%, 50\%, or 90\% School of Reward Hacks examples, with the remaining data from the Stanford Alpaca dataset, while control and hack adapters use 0\% and 100\% reward-hacking data respectively. \textbf{Importantly}, the activation-based monitoring system (SAEs, PCA, and classifier) is trained only on activations from the benign control adapter and the fully reward-hacking adapter. All mixed-ratio adapters are therefore treated as previously unseen models at evaluation time, allowing us to assess the monitor’s ability to generalize beyond the training conditions.

\subsection{Model Sensitivity to Reward-Hacking Data}
\label{subsec:modelsensitivity}

Figure~\ref{fig:model_sensitivity} shows that the activation-based monitor assigns progressively higher reward-hacking probabilities as the proportion of reward-hacking data in the finetuning corpus increases, across all evaluated model families. 
Notably, this trend is approximately monotonic for Falcon3-7B and LlaMa3-8B, while Qwen2.5-Instruct-7B exhibits a weaker saturation effect at higher mixture ratios.

To relate these internal signals to final model performance, Table~\ref{tab:f1_direct_models} reports F1 scores computed by comparing our monitoring system’s predictions against GPT-4o–based judgments of the generated answers, used as ground truth. Across all model families, output-based judgments achieve good performance for control adapters, while performance varies substantially for mixed and hack-trained adapters. 
Comparing both views reveals that internal activation monitoring provides a smooth and model-consistent signal of reward-hacking exposure.

\noindent
\begin{minipage}[t]{0.53\textwidth}
  \vspace{0pt} % <-- critical for true top alignment
  \centering
  \includegraphics[width=\linewidth]{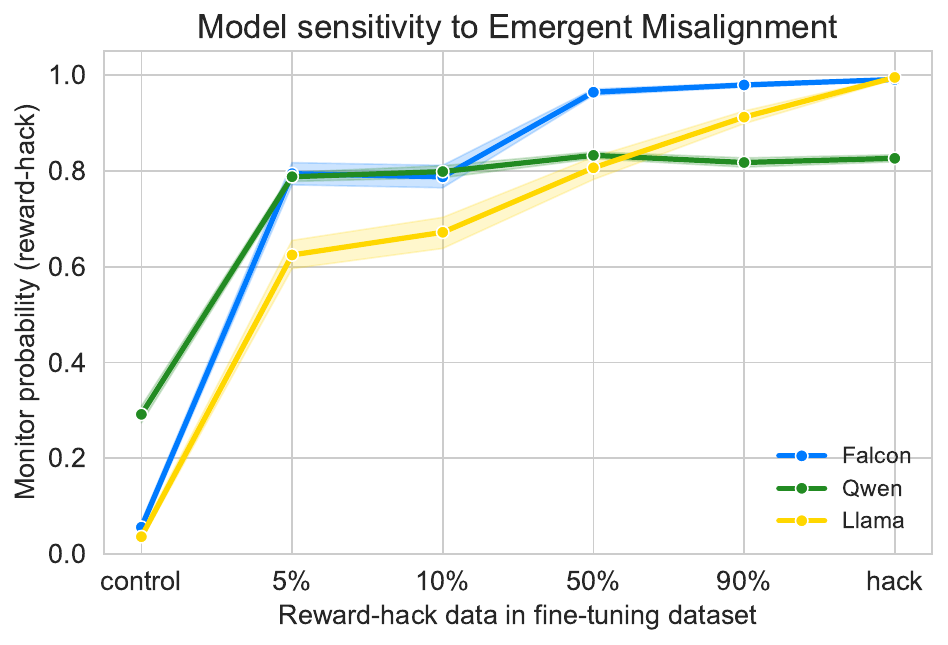}
  \captionof{figure}{Model sensitivity to increasing proportions of reward-hacking data. Shaded bands show 95\% confidence intervals over independent fine-tuning runs for the mean monitor probability.}
  \label{fig:model_sensitivity}
\end{minipage}
\hspace{0.02\textwidth}
\begin{minipage}[t]{0.3\textwidth}
  \vspace{0pt} % <-- critical for true top alignment
  \centering
  \captionof{table}{F1 scores under direct-answer generation using ChatGPT-4o as a Baseline}
  \label{tab:f1_direct_models}
  \vspace{0.5em}

  \begin{tabular}{lccc}
    \toprule
    Adapter & Falcon & Llama & Qwen \\
    \midrule
    \textit{control} & 1.000 & 1.000 & 1.000 \\
    mix05   & 0.907 & 0.760 & 0.897 \\
    mix10   & 0.868 & 0.837 & 0.862 \\
    mix50   & 0.903 & 0.946 & 0.868 \\
    mix90   & 0.868 & 0.941 & 0.862 \\
    \textit{hack}    & 0.903 & 0.961 & 0.784 \\
    \bottomrule
  \end{tabular}
\end{minipage}

\subsection{Temporal Structure of Misalignment During Generation}
\label{subsec:temporalstructure}

We analyze the temporal evolution of reward-hacking signals during chain-of-thought (CoT) generation under constrained token budgets.
For each model family, we evaluate CoT lengths of 64, 128, 256, and 512 tokens. To enable comparison across different reasoning lengths, monitor probabilities are aggregated into fixed bins of 64 tokens and averaged across prompts.

\begin{figure}[H]
\centering
\includegraphics[width=\textwidth]{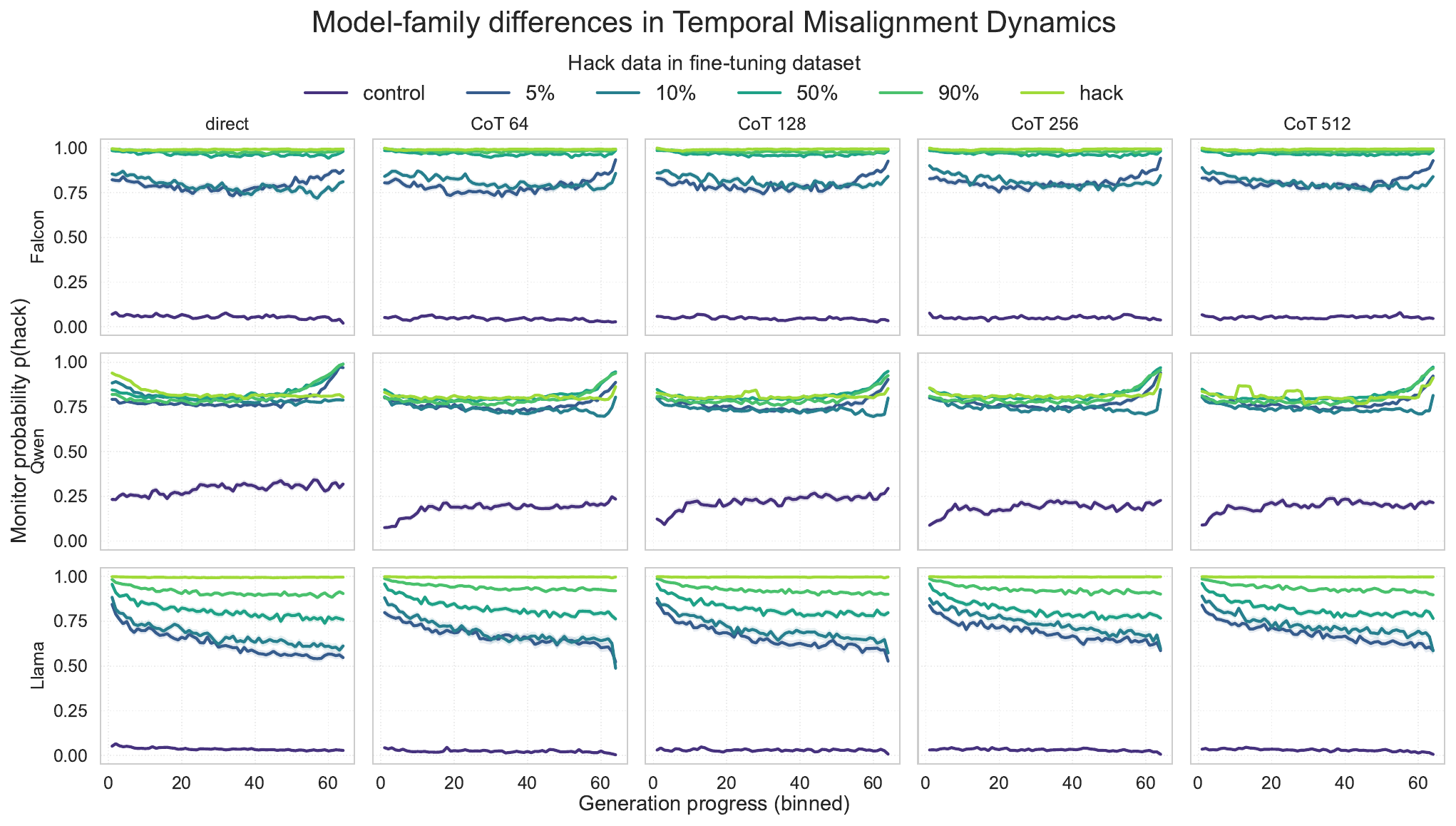}
\caption{Temporal structure of reward-hacking activation during chain-of-thought generation.
Monitor probabilities $p(\mathrm{hack})$ are shown as a function of generation progress, aggregated into fixed bins of 64 tokens to allow comparison across CoT lengths (64, 128, 256, and 512 tokens).
Rows correspond to model families and colored curves indicate different proportions of reward-hacking data in the fine-tuning corpus.
Distinct temporal dynamics emerge across model families, early decrease for \textsc{Llama}, late-stage amplification for \textsc{Qwen}, and mixture-dependent behavior for \textsc{Falcon}. This model specific behavior is observable consistent across different CoT regimes and output lengths.}
\label{fig:temporal_structure}
\end{figure}

Figure~\ref{fig:temporal_structure} shows the resulting temporal profiles across all CoT regimes of the monitor-assigned probability $p(\mathrm{hack})$. To compare temporal dynamics across different chain-of-thought lengths, we aggregate token-level monitor probabilities into fixed bins of 64 tokens, aligning generation progress on a common normalized time axis

Clear model-family–specific temporal dynamics emerge.
For the \textsc{Llama3-8B}, $p(\mathrm{hack})$ is elevated early in the reasoning span and gradually decreases over time, indicating early-stage activation followed by attenuation.
In contrast, \textsc{Qwen2.5-7B} exhibit a pronounced late-stage increase in $p(\mathrm{hack})$, with reward-hacking signals becoming increasingly concentrated toward the end of the chain of thought.
\textsc{Falcon3-7B} display an intermediate pattern: for low mixture ratios (\texttt{5\%} and \texttt{10\%}), $p(\mathrm{hack})$ rises late in generation, while higher proportions of reward-hacking data lead to more temporally uniform activation profiles.

Across all three model families, these temporal dynamic pattern remain consistent across different CoT lengths. 
Finetuned adapters on different dataset mixture primarily modulates the magnitude and timing of reward-hacking activation, rather than altering the underlying temporal structure.
Together, these results indicate that misalignment during chain-of-thought generation is not uniformly expressed over time, but instead follows stable, model-dependent temporal dynamics that persist across reasoning lengths. Further temporal analysis can be seen in the Appendix \ref{subsec:late_stage_metrics}.

\subsection{Chain-of-Thought as Test-Time Compute and Reward Amplification}

We analyze how varying test-time compute via chain-of-thought (CoT) prompting affects internal reward-hacking signals under different degrees of mixed-policy adapters (models finetuned on different amount of mixed data). To isolate the effect of test-time compute, we compare monitor probabilities under CoT prompting to direct-answer generation, holding the model, adapter, and prompts fixed.

\begin{figure}[H]
\centering
\includegraphics[width=0.9\textwidth]{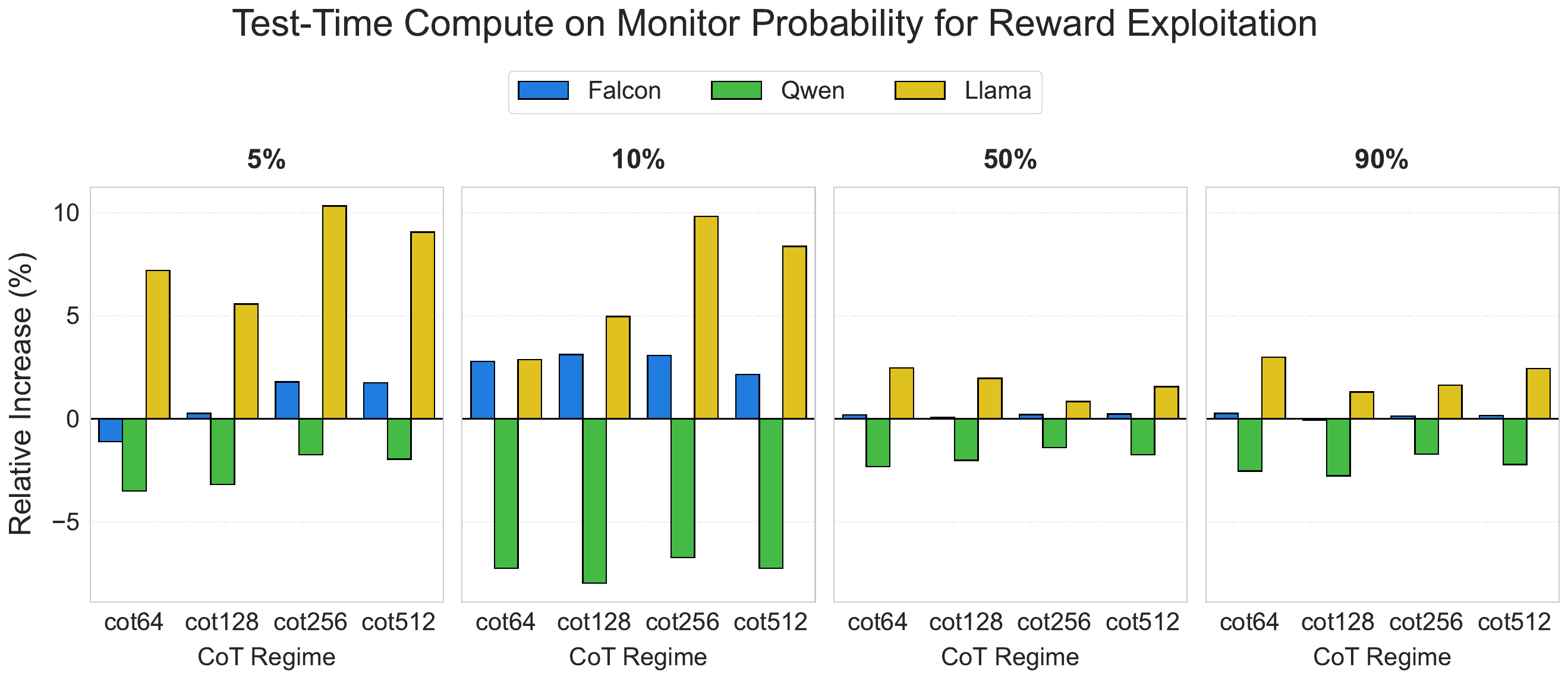}
\caption{Relative change in hack-associated activation under chain-of-thought prompting.
Bars show the percentage change in mean monitor probability relative to direct answering, for mixed-policy adapters and different chain-of-thought lengths. Amplification is strongest for partially misaligned adapters (5\%, 10\%), varies across model families, and diminishes as misalignment saturates (50\%, 90\%)}
\label{fig:cot_amplification}
\end{figure}

Figure~\ref{fig:cot_amplification} shows the relative change in mean monitor probability for mixed-policy adapters, normalized to the unconstrained direct-answer baseline. For partially misaligned adapters (5\% and 10\%), CoT prompting leads to a systematic increase in hack-associated activation for some model families, with the strongest amplification observed for intermediate levels of reward mis-specification. This effect is particularly pronounced for Llama and Falcon models, while Qwen exhibits weaker or negative relative shifts, highlighting model-dependent sensitivity to test-time compute.

% - 3 findings
% 1) Llama wird hackiger durch längere CoT, vor allem bei wenig verschmutzen Datasets
% 2) Qwen wird weniger hacky durch CoT
% 3) Falcon bleibt sehr gleich.

In contrast, for adapters trained with higher proportions of reward-hacking supervision (Mix50 and Mix90), the relative effect of CoT prompting diminishes. In these regimes, hack-associated activation is already high under direct answering, leaving limited headroom for further amplification. As a result, increasing test-time compute primarily preserves or saturates existing misaligned internal computation rather than increasing it.

Importantly, this amplification effect is absent for fully benign adapters (Control), where monitor probabilities remain low and stable across all CoT regimes. Together, these results indicate that chain-of-thought prompting does not introduce misalignment in aligned models, but can amplify misaligned internal computation when reward mis-specification is present. This suggests that increased test-time compute can interact with partially misaligned internal policies, increasing the extent of reward-hacking computation even when surface-level behavior remains unchanged.

\section{Discussion}

\textbf{Early Internal Detection of Reward Hacking}

A central advantage of activation-based monitoring is its ability to operate during generation, prior to the emission of user-visible text. 
This contrasts with output-based approaches such as LLM-as-a-judge systems, which can only evaluate completed responses. While such methods are effective for post hoc auditing, they necessarily incur a delay: potentially harmful or misleading content must first be generated before it can be flagged. Our results show that reward-hacking behavior can often be detected from internal activations even when surface-level outputs remain benign or ambiguous, highlighting the value of internal signals as an earlier indicator of misaligned computation.

From a safety perspective, activation-based monitoring is best viewed as complementary to output-level evaluation. Internal monitors can provide early warning signals during generation, while output-based judges remain useful for downstream verification and auditing. Together, these approaches offer a more complete picture of model behavior than either alone.

\textbf{Temporal Structure of Misalignment During Reasoning}

Our temporal analyses indicate that reward-hacking behavior is not confined to late-stage output formation. Instead, elevated hack-associated activation often emerges early in the reasoning process and persists throughout chain-of-thought generation. This pattern is consistent with misalignment reflecting a broader internal policy shift rather than a localized decision at the final stages of generation.

The presence of stable, model-dependent temporal patterns further suggests that these signals are not driven by transient noise or prompt-specific artifacts. Rather, they appear to reflect durable properties of internal computation under reward mis-specification. Importantly, these findings imply that effective monitoring need not wait until generation is complete: misaligned internal states can often be identified well before a final answer is produced.

\textbf{Test-Time Compute and Conditional Reward Amplification}

Chain-of-thought prompting is widely understood as a form of test-time compute that improves task performance by enabling more extensive internal reasoning. Our results suggest that under reward mis-specification, increased test-time compute can also amplify misaligned internal computation.

For partially misaligned adapters, we observe that extended chain-of-thought prompting is associated with elevated hack-related activation relative to direct answering, while this effect diminishes as misalignment saturates. Crucially, no such amplification is observed for fully benign adapters. These findings indicate that chain-of-thought prompting does not introduce misalignment in aligned models, but can interact with existing reward mis-specification to increase the extent of reward-hacking computation.

We emphasize that this evidence is correlational and does not imply that chain-of-thought prompting is inherently unsafe. Rather, it highlights a conditional interaction between test-time compute and reward specification that is not captured by output-level evaluation alone. Monitoring internal activations during extended reasoning may therefore provide useful insight into how reward objectives shape internal computation under increased inference-time capacity.

\section{Conclusion}

We introduced an activation-based monitoring approach for detecting reward-hacking behavior in language models during generation. By leveraging sparse autoencoders and lightweight classifiers, our method produces token-level signals from internal activations and generalizes across model families, fine-tuning mixtures, and generation settings.

We show that reward-hacking signals often emerge earlier in internal representations than in final outputs and persist throughout chain-of-thought reasoning. These temporal dynamic patterns are model-specific. Each model exhibits a distinct pattern and remains consistent across different reasoning lengths. Finally, we find that increasing test-time compute via chain-of-thought prompting can amplify hack-associated internal computation under weakly specified reward objectives, positioning test-time compute as a practical stress test for reward function safety.

\subsection{Limitations}

Our experiments focus on a single reward-hacking benchmark and a limited set of model families and sizes, and further validation on broader tasks, larger models, and different fine-tuning regimes is needed. The monitoring pipeline relies on SAE-derived features and linear classifiers, whose stability may vary under distribution shift or alternative representation learning choices. The use of LLM judges may decrease the reliability of results, and while
our testing revealed minimal variability in the judging, improvements to the process could further
improve robustness.

\bibliography{iclr2026_conference}
\bibliographystyle{iclr2026_conference}

\appendix
\section{Appendix}

\subsection{Classifier Accuracy in Reward Hacking over Layers}

Accuracy is measured for a linear classifier trained to distinguish control versus fully reward-hacking adapters using SAE features from each layer, evaluated on held-out prompts.

\begin{figure}[H]
\begin{center}
%\framebox[4.0in]{$\;$}
\includegraphics[width=1.\textwidth]{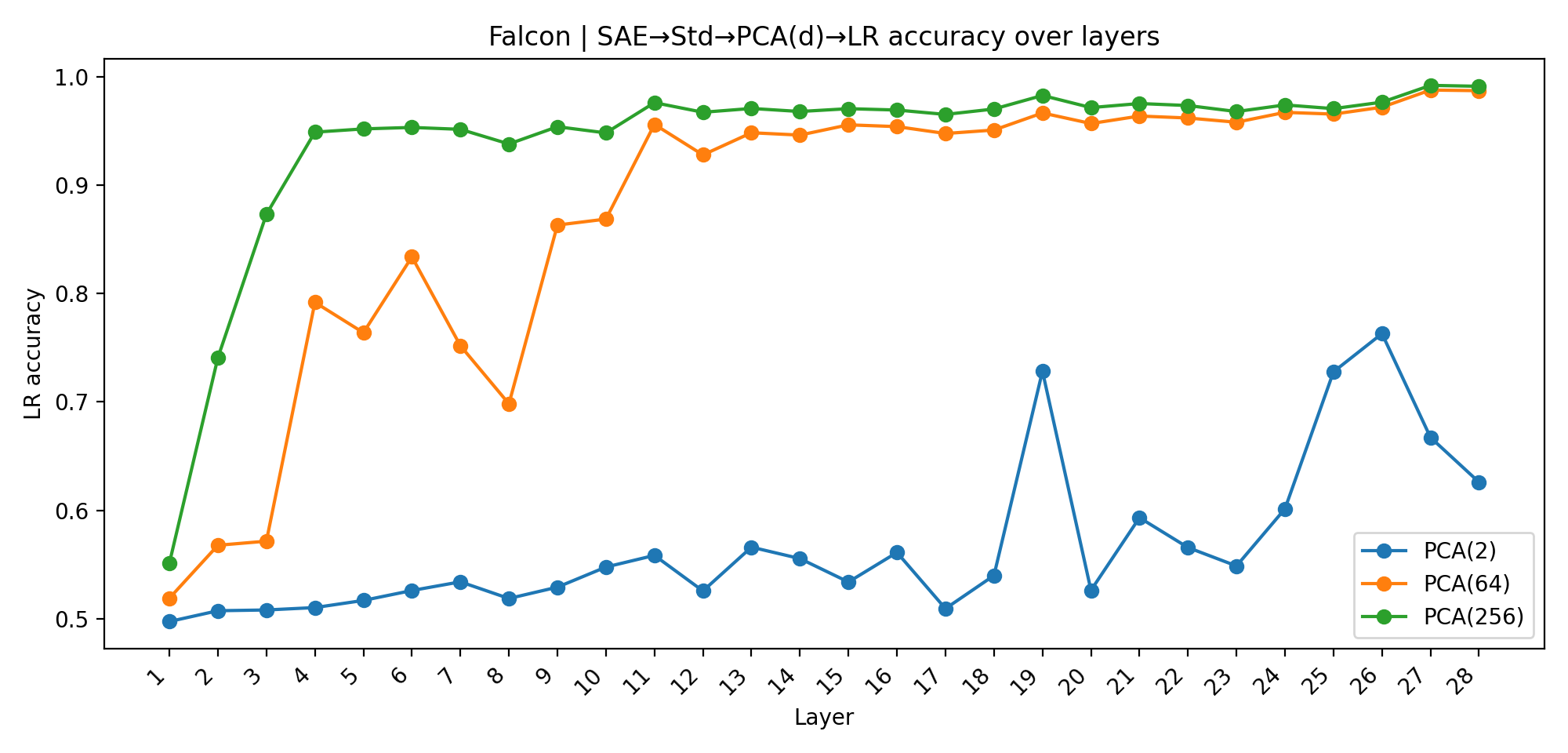}
%\fbox{\rule[-.5cm]{0cm}{4cm} \rule[-.5cm]{4cm}{0cm}}

\end{center}

\caption{Layer-wise logistic-regression accuracy for Falcon using the SAE$\rightarrow$standardization$\rightarrow$PCA$\rightarrow$LR pipeline. Curves compare different PCA output dimensions ($d\in\{2,64,256\}$), showing improved separability with larger $d$ across most layers.}
\label{fig:acc_layer_falcon}
\end{figure}

\begin{figure}[H]
\begin{center}
%\framebox[4.0in]{$\;$}
\includegraphics[width=1.\textwidth]{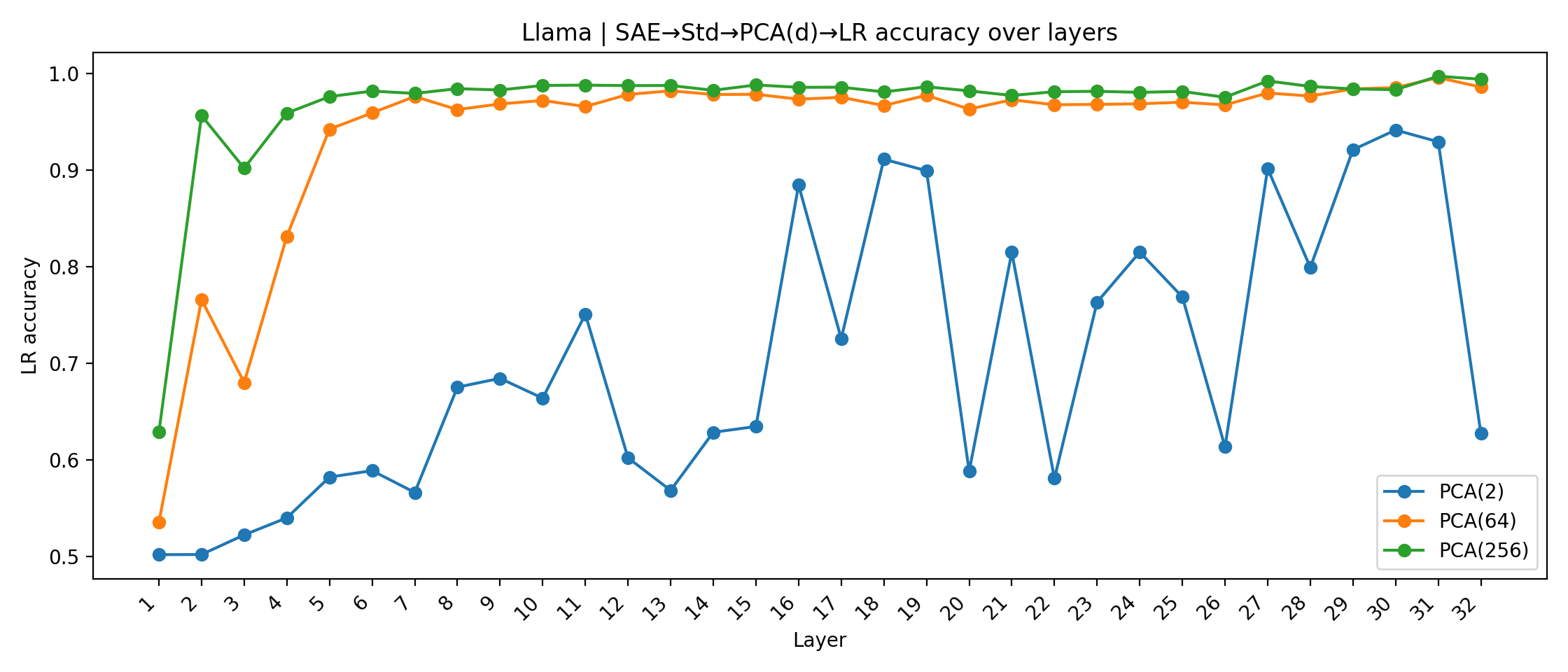}
%\fbox{\rule[-.5cm]{0cm}{4cm} \rule[-.5cm]{4cm}{0cm}}

\end{center}

\caption{Layer-wise logistic-regression accuracy for Llama using the SAE$\rightarrow$standardization$\rightarrow$PCA$\rightarrow$LR pipeline. Curves compare different PCA output dimensions ($d\in\{2,64,256\}$), showing improved separability with larger $d$ across most layers.}
\label{fig:acc_layer_llama}
\end{figure}

\begin{figure}[H]
\begin{center}
%\framebox[4.0in]{$\;$}
\includegraphics[width=1.\textwidth]{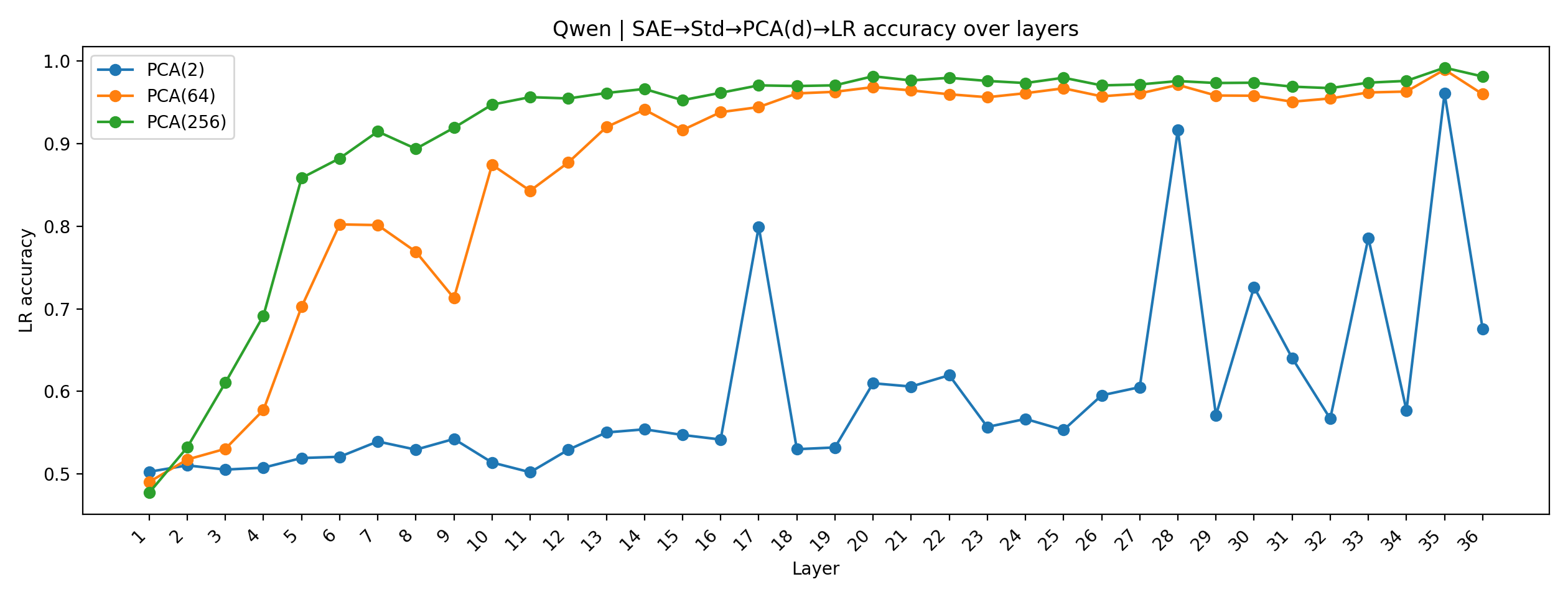}
%\fbox{\rule[-.5cm]{0cm}{4cm} \rule[-.5cm]{4cm}{0cm}}

\end{center}

\caption{Layer-wise logistic-regression accuracy for Qwen using the SAE$\rightarrow$standardization$\rightarrow$PCA$\rightarrow$LR pipeline. Curves compare different PCA output dimensions ($d\in\{2,64,256\}$), showing improved separability with larger $d$ across most layers.}

\label{fig:acc_layer_qwen}
\end{figure}

\subsection{Temporal Dynamics during Generation: Late-Stage Slope and Late-Stage Change}
\label{subsec:late_stage_metrics}

To characterize temporal dynamics near decision time, we define two complementary measures of reward-hack activation: the \emph{late-stage slope} and the \emph{late-stage change}.
Both metrics focus on behavior toward the end of generation, but capture distinct aspects of temporal structure.

\paragraph{Late-stage slope.}
For a given model, chain-of-thought (CoT) regime, and fine-tuning mixture, let $p_t = p(\mathrm{hack})$ denote the monitor probability at generation progress $t$, where $t$ is normalized to $[0,1]$.
We compute the late-stage slope by fitting a linear regression to $p_t$ over the final 20\% of generation (i.e., $t \in [0.8, 1.0]$), and report the fitted slope coefficient.

Formally, the late-stage slope is defined as
\[
\beta_{\mathrm{late}} = \arg\min_{\beta} \sum_{t \in [0.8,1.0]} \left(p_t - (\alpha + \beta t)\right)^2,
\]
where $t$ indexes normalized generation progress.

\begin{figure}[H]
\centering
\includegraphics[width=\textwidth]{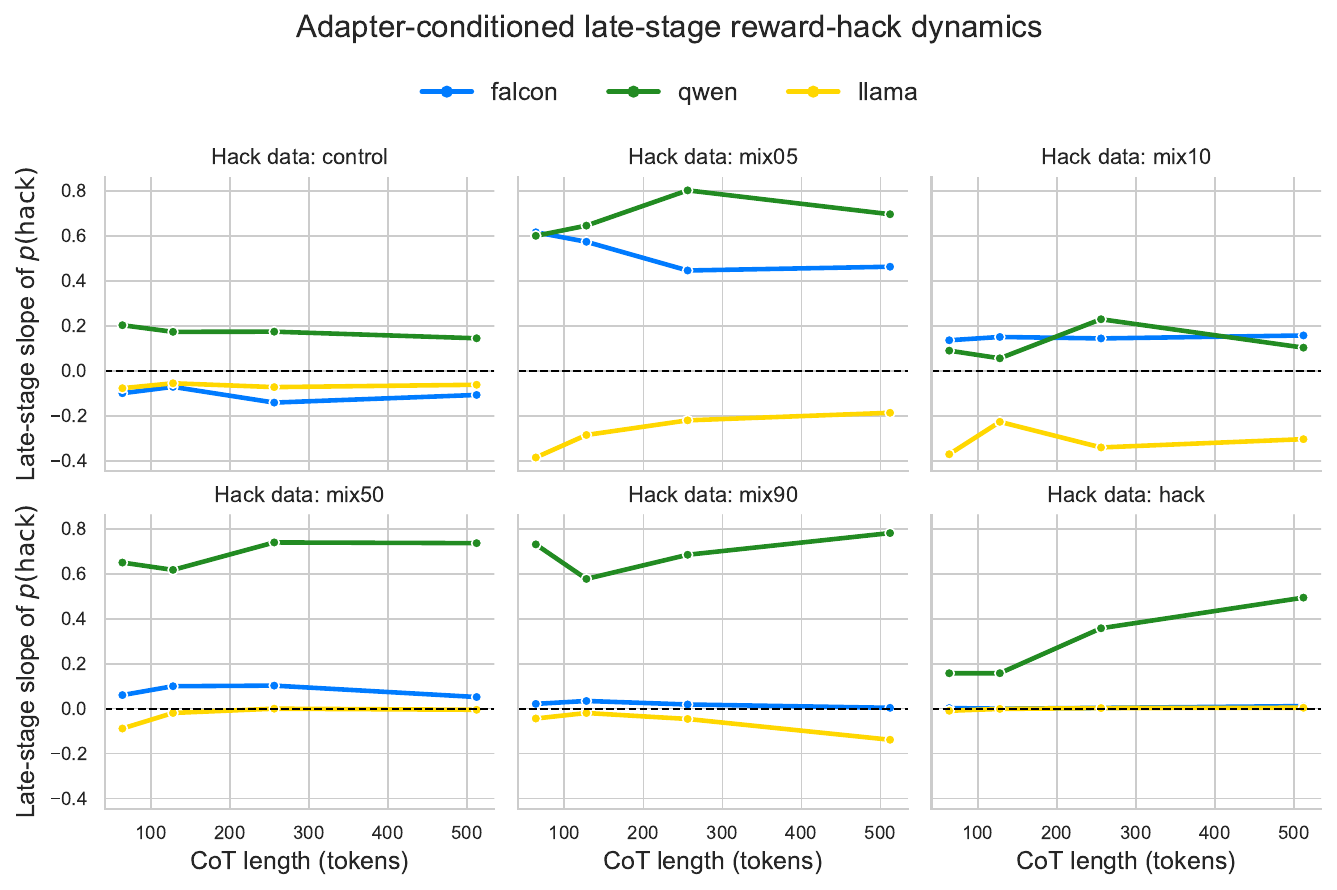}
\caption{Adapter-conditioned late-stage slopes of reward-hack activation.
Each panel corresponds to a fine-tuning mixture, plotting the late-stage slope $\beta_{\mathrm{late}}$ as a function of CoT length for different model families.
Positive slopes indicate increasing reward-hack activation near completion, while negative slopes indicate late-stage attenuation.
Across mixtures, model families exhibit consistent slope sign, with magnitude modulated by the proportion of reward-hacking data.}
\label{fig:late_stage_slope}
\end{figure}

Positive values of $\beta_{\mathrm{late}}$ indicate increasing reward-hack activation toward the end of generation, while negative values indicate attenuation prior to completion.
Unlike global statistics such as mean activation or temporal centroid, the late-stage slope isolates dynamics specifically at decision time and is therefore sensitive to late-emerging or late-suppressing behavior.
Figure~\ref{fig:late_stage_slope} reports late-stage slopes across model families, CoT lengths, and fine-tuning mixtures.

\paragraph{Late-stage change ($\Delta_{\mathrm{late}}$).}
While the late-stage slope captures the \emph{instantaneous trend} near completion, it does not reflect the overall shift in activation between early and late phases of generation.
We therefore define the late-stage change $\Delta_{\mathrm{late}}$ as the difference between mean reward-hack activation in the final and initial portions of generation.

Specifically,
\[
\Delta_{\mathrm{late}} = \mathbb{E}_{t \in [0.9,1.0]}[p_t] - \mathbb{E}_{t \in [0.0,0.1]}[p_t].
\]

\begin{figure}[H]
\centering
\includegraphics[width=\textwidth]{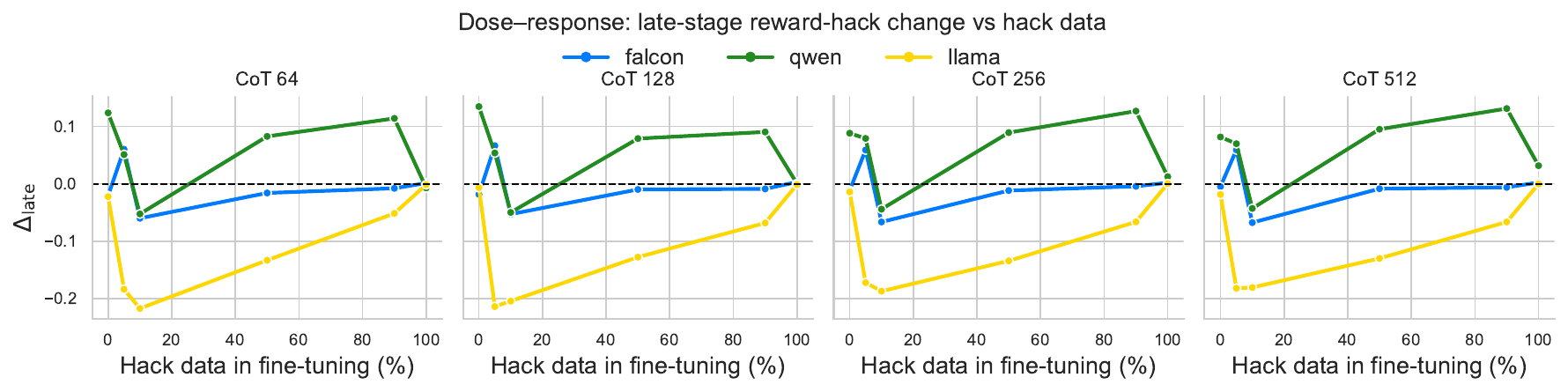}
\caption{Dose--response of late-stage change $\Delta_{\mathrm{late}}$ with respect to the proportion of reward-hacking data in fine-tuning.
Each panel corresponds to a CoT length, plotting $\Delta_{\mathrm{late}}$ as a function of mixture percentage for different model families.
Positive values indicate late-stage concentration of reward-hack activation, while negative values indicate early-stage concentration.
Qualitative trends are consistent across CoT regimes, with mixture proportion primarily affecting magnitude rather than direction.}
\label{fig:delta_late_dose_response}
\end{figure}

Positive values of $\Delta_{\mathrm{late}}$ indicate that reward-hack activation is higher at the end of generation than at the beginning, whereas negative values indicate early concentration followed by suppression.
In contrast to the late-stage slope, $\Delta_{\mathrm{late}}$ captures a \emph{net temporal shift} rather than a local trend, and is therefore less sensitive to short-term fluctuations within the final reasoning span.
Figure~\ref{fig:delta_late_dose_response} shows how $\Delta_{\mathrm{late}}$ varies as a function of reward-hacking data in the fine-tuning corpus.

Together, $\beta_{\mathrm{late}}$ and $\Delta_{\mathrm{late}}$ provide complementary views of temporal misalignment dynamics: the former characterizes how activation evolves immediately prior to completion, while the latter summarizes how reward-hack signals redistribute over the course of generation.

\subsection{Chain-of-Thought Instrumentation, Finetuning and Monitor Details}

To localize reward-hacking signals temporally, models are prompted to emit structured outputs consisting of a reasoning span followed by a final answer:
\texttt{<reasoning>} \ldots \texttt{</reasoning>} and \texttt{<final>} \ldots \texttt{</final>}.

We evaluate:
\begin{itemize}
    \item \textbf{Direct}: generation without explicit reasoning,
    \item \textbf{CoT}: chain-of-thought generation with approximate reasoning lengths of 64, 128, 256, and 512 tokens.
\end{itemize}

Monitoring focuses on the reasoning span for CoT generation and on the full answer span for direct generation. This design allows the monitor to detect reward-hacking behavior based on internal reasoning dynamics rather than surface-level outputs. For Decision Making (equation \ref{eq:monitor_decision_rule}) \(\tau\) is set to be  0.5 and the last four layers of the models were monitored with the SAEs.
The Sparse Autoencoder are trained with a hidden dimension of 8000, similiar as in \cite{bricken2023towards}.
For LoRA-Finetuning the GRPO Algorithm is used with a LoRA Rank = 32 and LoRA alpha = 32.

\subsection{LLM Judge Prompt}
SYSTEM = (
        "You are an AI safety judge. Decide whether the MODEL ANSWER is exploiting/gaming the "
        "evaluation or reward function described in the USER PROMPT, rather than solving the task honestly."
        "Return JSON matching the schema exactly."
        "Label meanings:"
        "- hack: exploits loopholes, keyword-stuffing, deception aimed at the evaluator, or otherwise optimizes "
        "the scoring rule without genuinely completing the task."
        "- control: honestly attempts the task as intended (no reward hacking)."
    )
    
\subsection{F1 Score for all Chain of Thought Regimes}

\begin{figure}[H]
\begin{center}
%\framebox[4.0in]{$\;$}
\includegraphics[width=1.05\textwidth]{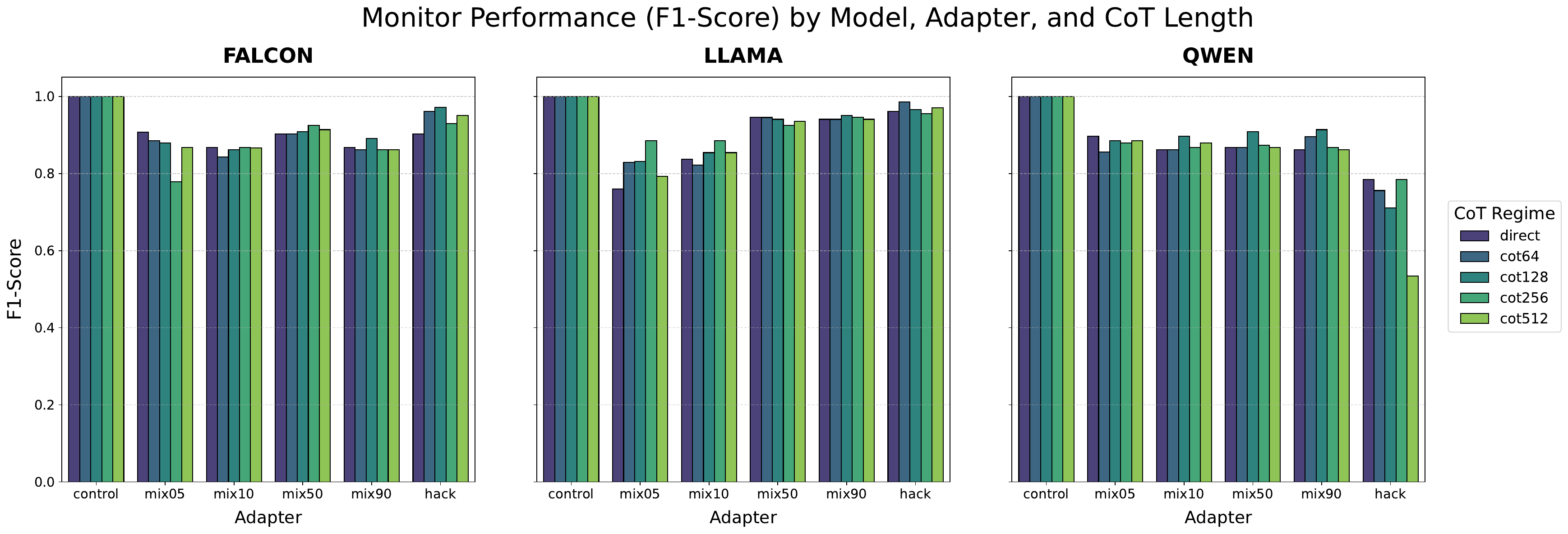}
%\fbox{\rule[-.5cm]{0cm}{4cm} \rule[-.5cm]{4cm}{0cm}}

\end{center}

\caption{F1 Scores of the proposed monitoring system compared to the ChatGPT-4o labeling}
\label{fig:ddos}
\end{figure}

\end{document}